\documentclass[runningheads]{llncs}
\usepackage[utf8]{inputenc}
\usepackage{graphicx}
\usepackage{amsmath,amssymb}
\usepackage{booktabs}
\usepackage{color}
\usepackage[table, dvipsnames]{xcolor}

\usepackage[width=122mm,left=12mm,paperwidth=146mm,height=193mm,top=12mm,paperheight=217mm]{geometry}

\usepackage[]{algorithm2e}
\usepackage{nicefrac}
\usepackage{subcaption}
\usepackage{adjustbox}
\usepackage{pifont}
\usepackage{multirow}

\usepackage[breaklinks=true,bookmarks=false]{hyperref}
\usepackage{placeins}
\usepackage[para,online,flushleft]{threeparttable}
\usepackage{nicefrac}

\begin{document}

\pagestyle{headings}
\mainmatter

\title{HybridNet: Classification and Reconstruction Cooperation for Semi-Supervised Learning}

\titlerunning{HybridNet: Classification and Reconstruction Cooperation for SSL}

\authorrunning{Thomas Robert, Nicolas Thome and Matthieu Cord}

\author{Thomas Robert\inst{1} \and Nicolas Thome\inst{2} \and Matthieu Cord\inst{1}}
\institute{
    Sorbonne Université, CNRS, LIP6, F-75005 Paris, France\\
    \email{\{thomas.robert,matthieu.cord\}@lip6.fr}
    \and
    CEDRIC - Conservatoire National des Arts et Métiers, 75003 Paris, France\\
    \email{nicolas.thome@cnam.fr}
}

\newcommand{\vx}{\mathbf{x}}
\newcommand{\vxh}{\hat{\mathbf{x}}}
\newcommand{\vy}{\mathbf{y}}
\newcommand{\vz}{\mathbf{z}}
\newcommand{\vyh}{\hat{\mathbf{y}}}
\newcommand{\vh}{\mathbf{h}}
\newcommand{\vhh}{\hat{\mathbf{h}}}
\newcommand{\calL}{\mathcal{L}}
\newcommand{\kk}{^{(k)}}

\maketitle

\begin{abstract}
In this paper, we introduce a new model for leveraging unlabeled data to improve generalization performances of image classifiers: a two-branch encoder-decoder architecture called HybridNet. The first branch receives supervision signal and is dedicated to the extraction of invariant class-related representations. The second branch is fully un\-supervised and dedicated to model information  discarded by the first branch to reconstruct input data. To further support the expected behavior of our model, we propose an original training objective. It favors stability in the discriminative branch and complementarity between the learned representations in the two branches. HybridNet is able to outperform state-of-the-art results on CIFAR-10, SVHN and STL-10 in various semi-supervised settings. In addition, visualizations and ablation studies validate our contributions and the behavior of the model on both CIFAR-10 and STL-10 datasets.

\end{abstract}

\section{Introduction}

Deep learning and Convolutional Neural Networks (ConvNets) have shown impressive state-of-the-art results in the last years on various visual recognition tasks, \textit{e.g.} image classification \cite{krizhevsky2012imagenet,he2016deep,Durand_WILDCAT_CVPR_2017}, object localization \cite{dai2016r,redmon2016you,Mordan2017}, image segmentation \cite{deeplab} and even multimodal embedding \cite{Martin2018,Carvalho2018,benyounescadene2017mutan}. Some key elements are the use of very deep models with a huge number of parameters and the availability of large-scale datasets such as ImageNet. When dealing with smaller datasets, however, the need for proper regularization methods becomes more crucial to control overfitting~\cite{weightdecay,batchnorm,dropout,Blot2018}. An appealing direction to tackle this issue is to take advantage of the huge number of unlabeled data by developing semi-supervised learning techniques.

Many approaches attempt at designing semi-supervised techniques where the unsupervised cost produces encoders that have high data-likelihood or small reconstruction error \cite{bengio2007greedy}. This strategy has been followed by historical deep learning approaches \cite{hinton2006reducing}, but also in some promising recent results with modern Conv\-Nets~\cite{Zhao2016a,Zhang2016a}. However, the unsupervised term in reconstruction-based approaches arguably conflicts with the supervised loss, which requires intra-class invariant representations. This is the motivation for designing auto-encoders that are able to discard information, such as the Ladder Networks~\cite{Rasmus2015}.

Another interesting regularization criterion relies on stability. Prediction functions which are stable under small input variations are likely to generalize well, especially when training with small amounts of data. Theoretical works have shown the stability properties of some deep models, \textit{e.g.} by using harmonic analysis for scattering transforms~\cite{Mallat2011,Bruna:2013:ISC:2498740.2498892} or for Convolution Kernel Machines~\cite{BiettiNIPS17}. In addition, recent semi-supervised models incorporate a stability-based regularizer on the prediction~\cite{Sajjadi2016,Laine2016,Tarvainen2017}.

\begin{figure}[tb]
	\centering
	\includegraphics[width=0.7\textwidth]{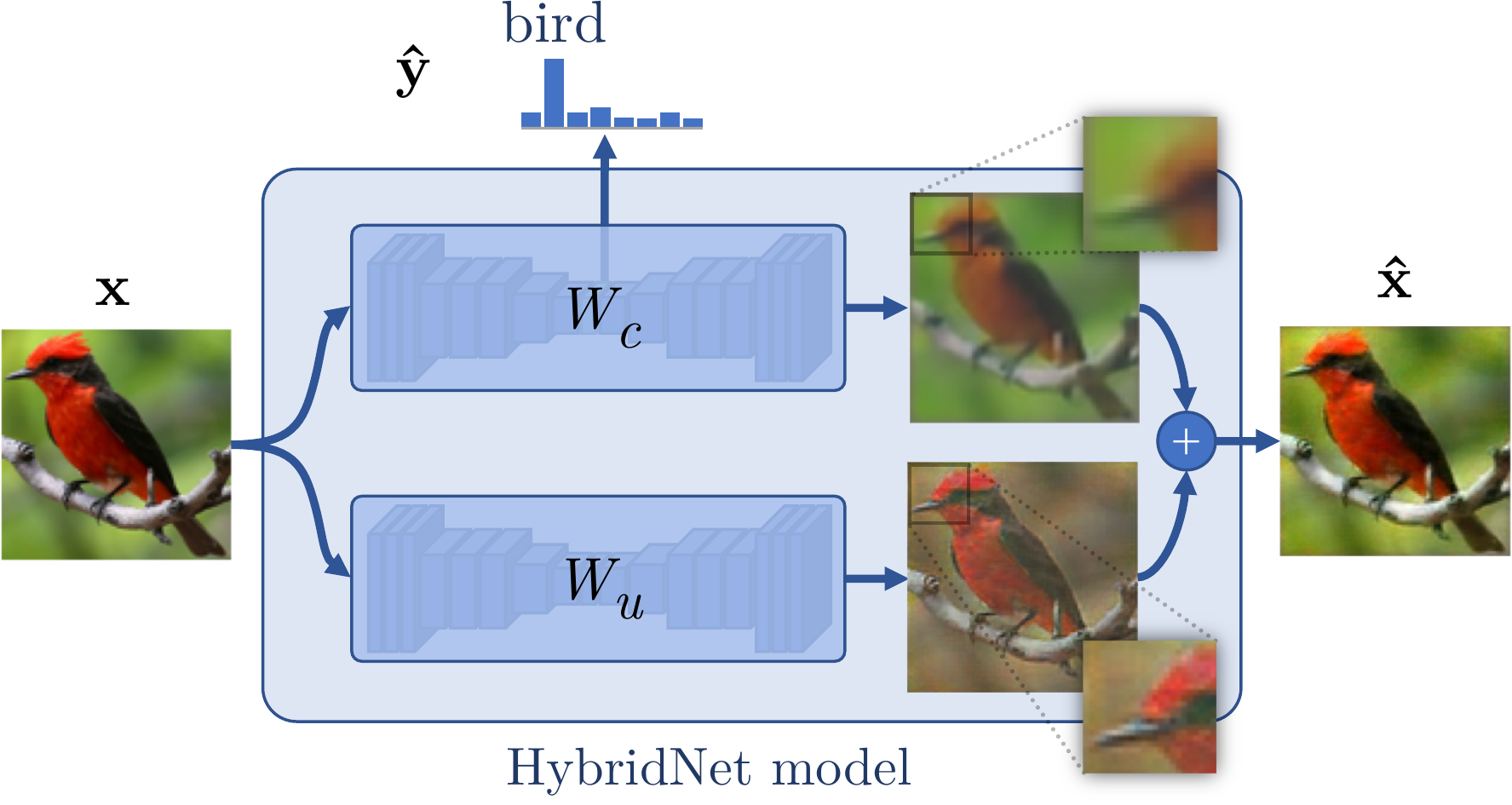}
    \caption{Illustration of HybridNet behavior: the input image is processed by two network paths of weights $W_c$ and $W_u$; each path produces a partial reconstruction, and both are summed to produce the final reconstruction, while only one path is used to produce a classification prediction. Thanks to a joint training of both tasks, the weights $W_c$ and $W_u$ influence each other to cooperate}
    \label{fig:intro}
\end{figure}

In this paper, we propose a new approach for regularizing ConvNets using unlabeled data. The behavior of our model, called HybridNet, is illustrated in Fig.~\ref{fig:intro}. It consists in a ``hybrid'' auto-encoder with the feature extraction path decomposed into two branches.

The top branch encoder, of parameters $W_c$, is connected to a classification layer that produces class predictions
while the decoder from this branch is used to partly reconstruct the input image from the discriminative features, leading to $\vxh_c$. Since those features are expected to extract invariant class-specific patterns, information is lost and exact reconstruction is not possible. To complement it, a second encoder-decoder branch of parameters $W_u$ is added to produce a complementary reconstruction $\vxh_u$ such that the sum $\vxh = \vxh_c+\vxh_u$ is the final complete reconstruction.

During training, the supervised classification cost impact $W_c$ while an unsupervised reconstruction cost is applied to both $W_c$ and $W_u$ to properly reconstruct the input image. The main assumption behind HybridNet is that the two-path architecture helps in making classification and reconstruction cooperate. To encourage this, we use additional costs and training techniques, namely a stability regularization in the discriminative branch and a branch complementarity training method.

\section{Related Work}

Training deep models with relatively small annotated datasets is a crucial issue nowadays.
To this end, the design of proper regularization techniques plays a central role.
In this paper, we address the problem of taking advantage of unlabeled data for improving generalization performances of deep ConvNets with semi-supervised learning~\cite{Zhu2005}.

One standard goal followed when training deep models with unlabeled data consists in designing models which fit input data well. Reconstruction error is the standard criterion used in (possibly denoising) Auto-Encoders~\cite{bengio2007greedy,Ranzato2008,Ranzato2007b,vincent2008extracting}, while maximum likelihood is used with generative models, \textit{e.g.} Restricted Boltzmann Machines, Deep Belief Networks or Deep Generative Models \cite{hinton2006reducing,Ranzato2007,Larochelle2008,kingma2014semi}. This unsupervised training framework was generally used as a pre-training before supervised learning with back-propagation \cite{erhan2010does}, potentially with an intermediate step \cite{Goh_NIPS13}. The currently very popular Generative Adversarial Networks~\cite{NIPS2014_5423} also falls into this category. With modern ConvNets, regularization with unlabeled data is generally formulated as a multi-task learning problem, where reconstruction and classification objectives are combined during training~\cite{Zhao2016a,Zhang2016a,Makhzani2016}. In these architectures, the encoder used for classification is regularized by a decoder dedicated to reconstruction.

This strategy of classification and reconstruction with an Auto-Encoder is however questionable, since classification and reconstruction may play contradictory roles in terms of feature extraction. Classification arguably aims at extracting invariant class-specific features, improving sample complexity of the learned model~\cite{hastie_09_elements-of.statistical-learning}, therefore inducing an information loss which prevents exact reconstruction. Ladder Networks~\cite{Rasmus2015} have historically been designed to overcome the previously mentioned conflict between reconstruction and classification, by designing Auto-Encoders capable of discarding information. Reconstruction is produced using higher-layer representation and a noisy version of the reconstruction target. However, it is not obvious that providing a noisy version of the target and training the network to remove the noise allows the encoder to lose some information since it must be able to correct low-level errors that require details.

Another interesting regularization criterion relies on stability or smoothness of the prediction function, which is at the basis of interesting unsupervised training methods, \textit{e.g.} Slow Feature Analysis~\cite{TheriaultCVPR13}. Adding stability to the prediction function was studied in Adversarial Training~\cite{goodfellow2014explaining} for supervised learning and further extended to semi-supervised learning in the Virtual Adversarial Training method \cite{miyato2015distributional}. Other recent semi-supervised models incorporate a stability-based regularizer on the prediction. The idea was first introduced by~\cite{Sajjadi2016} and proposes to make the prediction vector stable toward data augmentation (translation, rotation, shearing, noise, \textit{etc.}) and model stochasticity (dropout) for a given input. Following work~\cite{Laine2016,Tarvainen2017} slightly improves upon it by proposing variants on the way to compute stability targets to increase their consistency and better adapt to the model's evolution over training.

When using large modern ConvNets, the problem of designing decoders able to invert the encoding still is an open question~\cite{Wojna2017}. The usual solution is to mirror the architecture of the encoder by using transposed convolutions~\cite{dumoulin2016guide}.
This problem is exacerbated with irreversible pooling operations such as max-pooling that must be reversed by an upsampling operation. In~\cite{Zhao2016a,Zhang2016a}, they use unpooling operations to bring back spatial information from the encoder to the decoder, reusing pooling switches locations for upsampling.
Another interesting option is to explicitly create models which are reversible by design. This is the option followed by recent works such as RevNet~\cite{NIPS2017_6816} and i-RevNet~\cite{jacobsen:hal-01712808}, being inspired by second generation of bi-orthogonal multi-resolution analysis and wavelets~\cite{swe:spie95} from the signal processing literature.

To sum up, using reconstruction as a regularization cost added to classification is an appealing idea but the best way to efficiently use it as a regularizer is still an open question. As we have seen, when applied to an auto-encoding architecture~\cite{Zhao2016a,Zhang2016a}, reconstruction and classification would compete. To overcome the aforementioned issues, we propose HybridNet, a new framework for semi-supervised learning. Presented on Fig.~\ref{fig:general-archi}, this framework can be seen as an extension of the popular auto-encoding architecture.
In HybridNet, the usual auto-encoder that does both classification and reconstruction is assisted by an additional auto-encoder so that the first one is allowed to discard information in order to produce intra-class invariant features while the second one retains the lost information. The combination of both branches then produces the reconstruction.
This way, our architecture prevents the conflict between classification and reconstruction and allows the two branches to cooperate and accomplish both classification and reconstruction tasks.

Compared to Ladder Networks~\cite{Rasmus2015}, our two-branch approach without direct skip connection allows for a finer and learned information separation and is thus expected to have a more favorable impact in terms of discriminative encoder regularization. Our HybridNet model also has conceptual connections with wavelet decomposition~\cite{wavelets}: the first branch can be seen as extracting discriminative low-pass features from input images, and the second branch acting as a high-pass filter to restore the lost information. HybridNet also differs from reversible models~\cite{NIPS2017_6816,jacobsen:hal-01712808} by the explicit decomposition between supervised and unsupervised signals, enabling the discriminative encoder to have fewer parameters and better sample complexity.

In this paper, our contributions with the HybridNet framework are twofold: first, in Section~\ref{sec:archi}, we propose an architecture designed to efficiently allow both reconstruction and classification losses to cooperate; second, in Section~\ref{sec:training}, we design a training loss adapted to it that includes reconstruction, stability in the discriminative branch and a branch complementarity technique. In Section~\ref{sec:experiements}, we perform experiments to show that HybridNet is able to outperform state-of-the-art results in various semi-supervised settings on CIFAR-10, SVHN and STL-10. We also provide ablation studies validating the favorable impact of our contributions. Finally, we show several visualizations on CIFAR-10 and STL-10 datasets analogous to Fig.~\ref{fig:intro} to validate the behavior of both branches, with a discriminative branch that loses information that is restored by the second branch.

\section{HybridNet: a semi-supervised learning framework}
\label{sec:architecture}

In this section, we detail the proposed HybridNet model: the chosen architecture to mix supervised and unsupervised information efficiently in Section~\ref{sec:archi}, and the semi-supervised training method adapted to this particular architecture in Section~\ref{sec:training}.

\subsection{Designing the HybridNet architecture}
\label{sec:archi}

\begin{figure}[tb]
	\centering
	\includegraphics[width=0.7\textwidth]{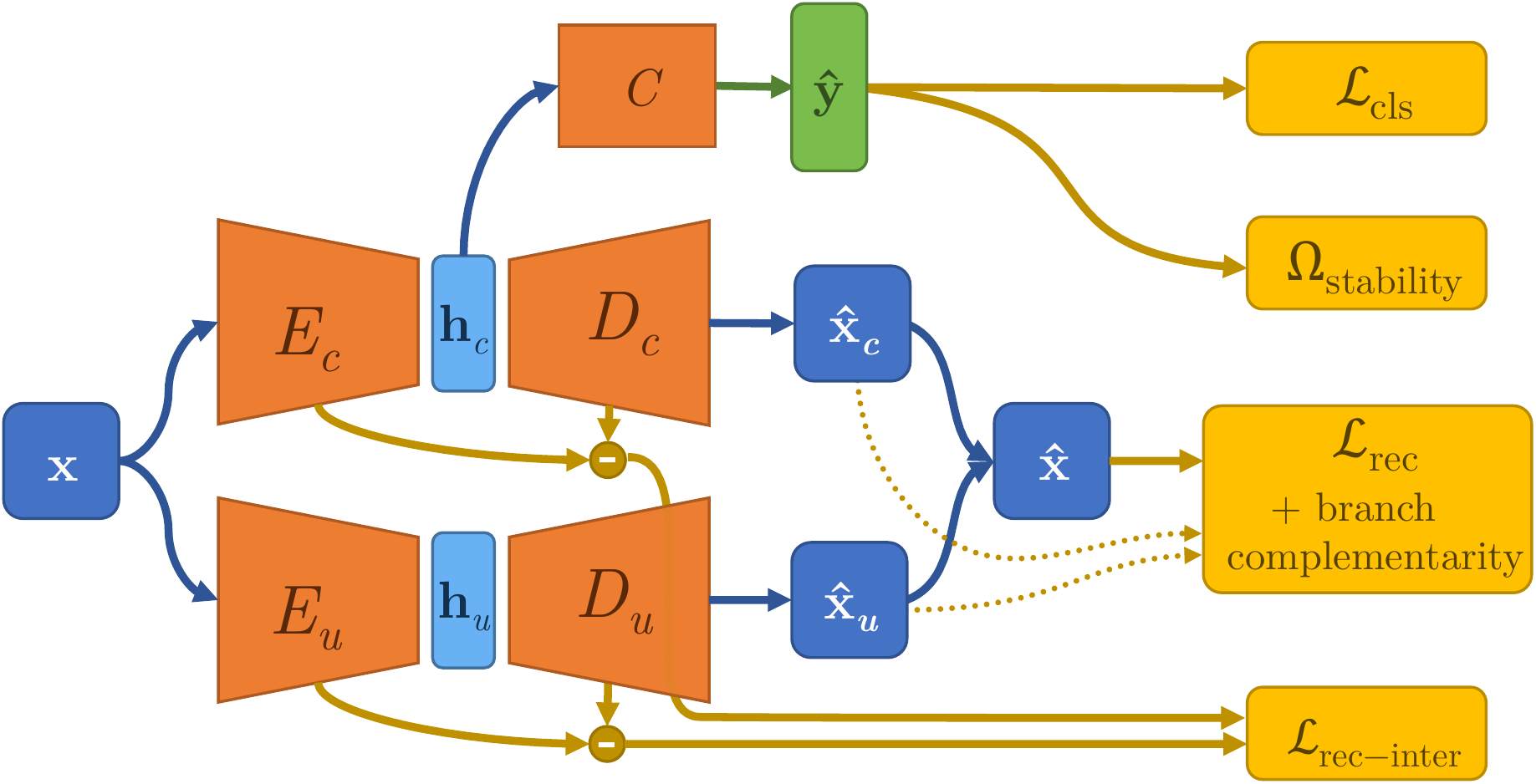}
    \caption{General description of the HybridNet framework. $E_c$ and $C$ correspond to a classifier, $E_c$ and $D_c$ form an autoencoder that we call \textit{discriminative path}, and $E_u$ and $D_u$ form a second autoencoder called \textit{unsupervised path}. The various loss functions used to train HybridNet are also represented in yellow}
    \label{fig:general-archi}
\end{figure}

\subsubsection{General architecture.}
As we have seen, classification requires intra-class invariant features while reconstruction needs to retain all the information. To circumvent this issue, HybridNet is composed of two auto-encoding paths, the \textit{discriminative path} ($E_c$ and $D_c$) and the \textit{unsupervised path} ($E_u$ and $D_u$). Both encoders $E_c$ and $E_u$ take an input image $\vx$ and produce representations $\vh_c$ and $\vh_u$, while decoders $D_c$ and $D_u$ take respectively $\vh_c$ and $\vh_u$ as input to produce two partial reconstructions $\vxh_c$ and $\vxh_u$. Finally, a classifier $C$ produces a class prediction using discriminative features only: $\vyh = C(\vh_c)$. Even if the two paths can have similar architectures, they should play different and complementary roles. The discriminative path must extract discriminative features $\vh_c$ that should eventually be well crafted to perform a classification task effectively, and produce a purposely partial reconstruction $\vxh_c$ that should not be perfect since preserving all the information is not a behavior we want to encourage. Consequently, the role of the unsupervised path is to be complementary to the discriminative branch by retaining in $\vh_u$ the information lost in $\vh_c$. This way, it can produce a complementary reconstruction $\vxh_u$ so that, when merging $\vxh_u$ and $\vxh_c$, the final reconstruction $\vxh$ is close to $\vx$. The HybridNet architecture, visible on Fig.~\ref{fig:general-archi}, can be described by the following equations:
\begin{equation}\arraycolsep=8pt
\begin{array}{ccc}
	\vh_c = E_c(\vx) & \vxh_c = D_c(\vh_c) & \vyh = C(\vh_c)\\
	\vh_u = E_u(\vx) & \vxh_u = D_u(\vh_u) & \vxh = \vxh_c + \vxh_u
\end{array}
\end{equation}

Note that the end-role of reconstruction is just to act as a regularizer for the discriminative encoder.
The main challenge and contribution of this paper is to find a way to ensure that the two paths will in fact behave in this desired way. The two main issues that we tackle are the fact that we want the discriminative branch to focus on discriminative features, and that we want both branches to cooperate and contribute to the reconstruction. Indeed, with such an architecture, we could end up with two paths that work independently: a classification path $\vyh = C(E_c(\vx))$ and a reconstruction path $\vxh = \vxh_u = D_u(E_u(\vx))$ and $\vxh_c = 0$. We address both those issues through the design of the architecture of the encoders and decoders as well as an appropriate loss and training procedure.

\subsubsection{Branches design.}

To design the HybridNet architecture, we start with a convolutional architecture adapted to the targeted dataset, for example a state-of-the-art ResNet architecture for CIFAR-10. This architecture is split into two modules: the discriminative encoder $E_c$ and the classifier $C$. On top of this model, we add the discriminative decoder $D_c$.
The location of the splitting point in the original network is free, but $C$ will not be directly affected by the reconstruction loss. In our experiments, we choose $\vh_c$ ($E_c$'s output) to be the last intermediate representation before the final pooling that aggregates all the spatial information, leaving in $C$ a global average pooling followed by one or more fully-connected layers. The decoder $D_c$ is designed to be a ``mirror'' of the encoder's architecture, as commonly done in the literature, \textit{e.g.}~\cite{Zhao2016a,Rasmus2015,zeiler2014visualizing}.

After constructing the discriminative branch, we add an unsupervised complementary branch. To ensure that both branches are ``balanced'' and behave in a similar way, the internal architecture of $E_u$ and $D_u$ is mostly the same as for $E_c$ and $D_c$. The only difference remains in the mirroring of pooling layers, that can be reversed either by upsampling or unpooling. An upsampling will increase the spatial size of a feature map without any additional information while an unpooling, used in~\cite{Zhao2016a,Zhang2016a}, will use spatial information (\textit{pooling switches}) from the corresponding max-pooling layer to do the upsampling. In our architecture, we propose to use upsampling in the discriminative branch because we want to encourage spatial invariance, and use unpooling in the unsupervised branch to compensate this information loss and favor the learning of spatial-dependent low-level information. An example of HybridNet architecture is presented in Fig.~\ref{fig:cifar10-archi}.

As mentioned previously, one key problem to tackle is to ensure that this model will behave as expected, \textit{i.e.} by learning discriminative features in the discriminative encoder and non-discriminative features in the unsupervised one.
This is encouraged in different ways by the design of the architecture. First, the fact that only $\vh_c$ is used for classification means that $E_c$ will be pushed by the classification loss to produce discriminative features. Thus, the unsupervised branch will naturally focus on information lost by $E_c$. Using upsampling in $D_c$ and unpooling in $D_u$ also encourages the unsupervised branch to focus on low-level information. In addition to this, the design of an adapted loss and training protocol is a major contribution to the efficient training of HybridNet.

\subsection{Training HybridNet}
\label{sec:training}

The HybridNet architecture has two information paths with only one producing a class prediction and both producing partial reconstructions that should be combined. In this section, we will address the question of training this architecture efficiently. The complete loss is composed of various terms as illustrated on Fig.~\ref{fig:general-archi}. It comprises terms for classification with $\mathcal L_\textrm{cls}$; final reconstruction with ${\mathcal L_\textrm{rec}}$; intermediate reconstructions with ${\mathcal L_\textrm{rec-inter}}_{b,l}$ (for layer $l$ and branch $b$); and stability with $\mathrm{\Omega}_\textrm{stability}$. It is also accompanied by a branch complementarity training method. Each term is weighted by a corresponding parameter $\lambda$:
\begin{equation}\textstyle
	\mathcal L = \lambda_\textrm{c} \mathcal L_\textrm{cls} + \lambda_\textrm{r} \mathcal L_\textrm{rec} + \sum_{b\in \{c,u\},l} {\lambda_\textrm{r}}_{b,l} {\mathcal L_\textrm{rec-inter}}_{b,l} + \lambda_\textrm{s} \mathrm{\Omega}_\textrm{stability} \,.
    \label{eq:full-loss}
\end{equation}

HybridNet can be trained on a partially labeled dataset, \textit{i.e.} that is composed of labeled pairs $\mathcal D_\textrm{sup} = \{(x^{(k)}, y^{(k)})\}_{k=1..N_\textrm{s}}$ and unlabeled images $\mathcal D_\textrm{unsup} = \{x^{(k)}\}_{k=1..N_\textrm{u}}$.
Each batch is composed of $n$ samples, divided into $n_\textrm{s}$ image-label pairs from $\mathcal D_\textrm{sup}$ and $n_\textrm{u}$ unlabeled images from $\mathcal D_\textrm{unsup}$.

\subsubsection{Classification.}

The classification term is a regular cross-entropy term, that is applied only on the $n_s$ labeled samples of the batch and averaged over them:
\begin{equation}
	\ell_\mathrm{cls} = \ell_\mathrm{CE}(\vyh, \vy) = -\sum_i \vy_i \log \vyh_i \, , \qquad \mathcal L_\textrm{cls}=\frac{1}{n_s} \sum_k \ell_\mathrm{cls}(\vyh\kk, \vy\kk) \ .
\end{equation}

\subsubsection{Reconstruction losses.}

In HybridNet, we chose to keep discriminative and unsupervised paths separate so that they produce two complementary reconstructions $(\vxh_u, \vxh_c)$ that we combine with an addition into $\vxh = \vxh_u + \vxh_c$. Keeping the two paths independent until the reconstruction in pixel space, as well as the merge-by-addition strategy allows us to apply different treatments to them and influence their behavior efficiently. The merge by addition in pixel space is also analogous to wavelet decomposition where the signal is decomposed into low- and high-pass branches that are then decoded and summed in pixel space. The reconstruction loss that we use is a simple mean-squared error between the input and the sum of the partial reconstructions:
\begin{equation}
	\ell_\mathrm{rec} = ||\vxh - \vx||_2^2 = ||\vxh_u + \vxh_c - \vx||_2^2\,, \qquad \mathcal L_\textrm{rec}=\frac{1}{n} \sum_k \ell_\mathrm{rec}(\vxh\kk, \vx\kk) \ .
\end{equation}

In addition to the final reconstruction loss, we also add reconstruction costs between intermediate representations in the encoders and the decoders which is possible since encoders and decoders have mirrored structure. We apply these costs to the representations $\vh_{b,l}$ (for branch $b$ and layer $l$) produced just after pooling layers in the encoders and reconstructions $\vhh_{b,l}$ produced just before the corresponding upsampling or unpooling layers in the decoders. This is common in the literature \cite{Zhao2016a,Zhang2016a,Rasmus2015} but is particularly important in our case: in addition to guiding the model to produce the right final reconstruction, it pushes the discriminative branch to produce a reconstruction and avoid the undesired situation where only the unsupervised branch would contribute to the final reconstruction. This is applied in both branches ($b \in \{c,u\}$):
\begin{equation}
	{\mathcal L_\textrm{rec-inter}}_{b,l} = \frac{1}{n} \sum_k ||\vhh_{b,l}\kk - \vh_{b,l}\kk||_2^2\,.
\end{equation}

\subsubsection{Branch cooperation.} As described previously, we want to ensure that both branches contribute to the final reconstruction, otherwise this would mean that the reconstruction is not helping to regularize $E_c$, which is our end-goal. Having both branches produce a partial reconstruction and using intermediate reconstructions already help with this goal. In addition, to balance their training even more, we propose a training technique such that the reconstruction loss is only backpropagated to the branch that contributes less to the final reconstruction of each sample. This is done by comparing $||\vxh_c - \vx||_2^2$ and $||\vxh_u - \vx||_2^2$ and only applying the final reconstruction loss to the branch with the higher error.

This can be implemented either in the gradient descent or simply by preventing gradient propagation in one branch or the other using features like \texttt{tf.stop\_gradient} in Tensorflow or \texttt{.detach()} in PyTorch:
\begin{equation}
	\mathcal \ell_\textrm{rec-balanced} = \begin{cases}
    	||\vxh_u + \mathrm{stopgrad}(\vxh_c) - \vx||_2^2&\text{if }||\vxh_u - \vx||_2^2 \geq ||\vxh_c - \vx||_2^2\\
        ||\mathrm{stopgrad}(\vxh_u) + \vxh_c - \vx||_2^2&\text{otherwise}\\
    \end{cases} .
\end{equation}

\subsubsection{Encouraging invariance in the discriminative branch.}

We have seen that an important issue that needs to be addressed when training this model is to ensure that the discriminative branch will filter out information and learn invariant features. For now, the only signal that pushes the model to do so is the classification loss. However, in a semi-supervised context, when only a small portion of our dataset is labeled, this signal can be fairly weak and might not be sufficient to make the discriminative encoder focus on invariant features.

In order to further encourage this behavior, we propose to use a \textit{stability regularizer}. Such a regularizer is currently at the core of the models that give state-of-the-art results in semi-supervised setting on the most common datasets~\cite{Sajjadi2016,Laine2016,Tarvainen2017}. The principle is to encourage the classifier's output prediction $\vyh\kk$ for sample $k$ to be invariant to different sources of randomness applied on the input (translation, horizontal flip, random noise, \textit{etc.}) and in the network (\textit{e.g.} dropout). This is done by minimizing the squared euclidean distance between the output $\vyh\kk$ and a ``stability'' target $\vz\kk$. Multiple methods have been proposed to compute such a target~\cite{Sajjadi2016,Laine2016,Tarvainen2017}, for example by using a second pass of the sample in the network with a different draw of random factors that will therefore produce a different output. We have:
\begin{equation}
	\mathrm{\Omega}_\mathrm{stability} = \frac{1}{n} \sum_k ||\vyh\kk - \vz\kk||_2^2\,.
\end{equation}

By applying this loss on $\vyh$, we encourage $E_c$ to find invariant patterns in the data, patterns that have more chances of being discriminative and useful for classification. Furthermore, this loss has the advantage of being applicable to both labeled and unlabeled images.

In the experiments, we tried both Temporal Ensembling~\cite{Laine2016} and Mean Teacher~\cite{Tarvainen2017} methods and did not see a major difference. In Temporal Ensembling, the target $\vz\kk$ is a moving average of the $\vyh\kk$ over the previous pass of $\vx\kk$ in the network during training; while in Mean Teacher, $\vz\kk$ is the output of a secondary model where weights are a moving average of the weights of the model being trained.

\begin{figure}[tb]
	\centering
	\includegraphics[width=0.75\textwidth]{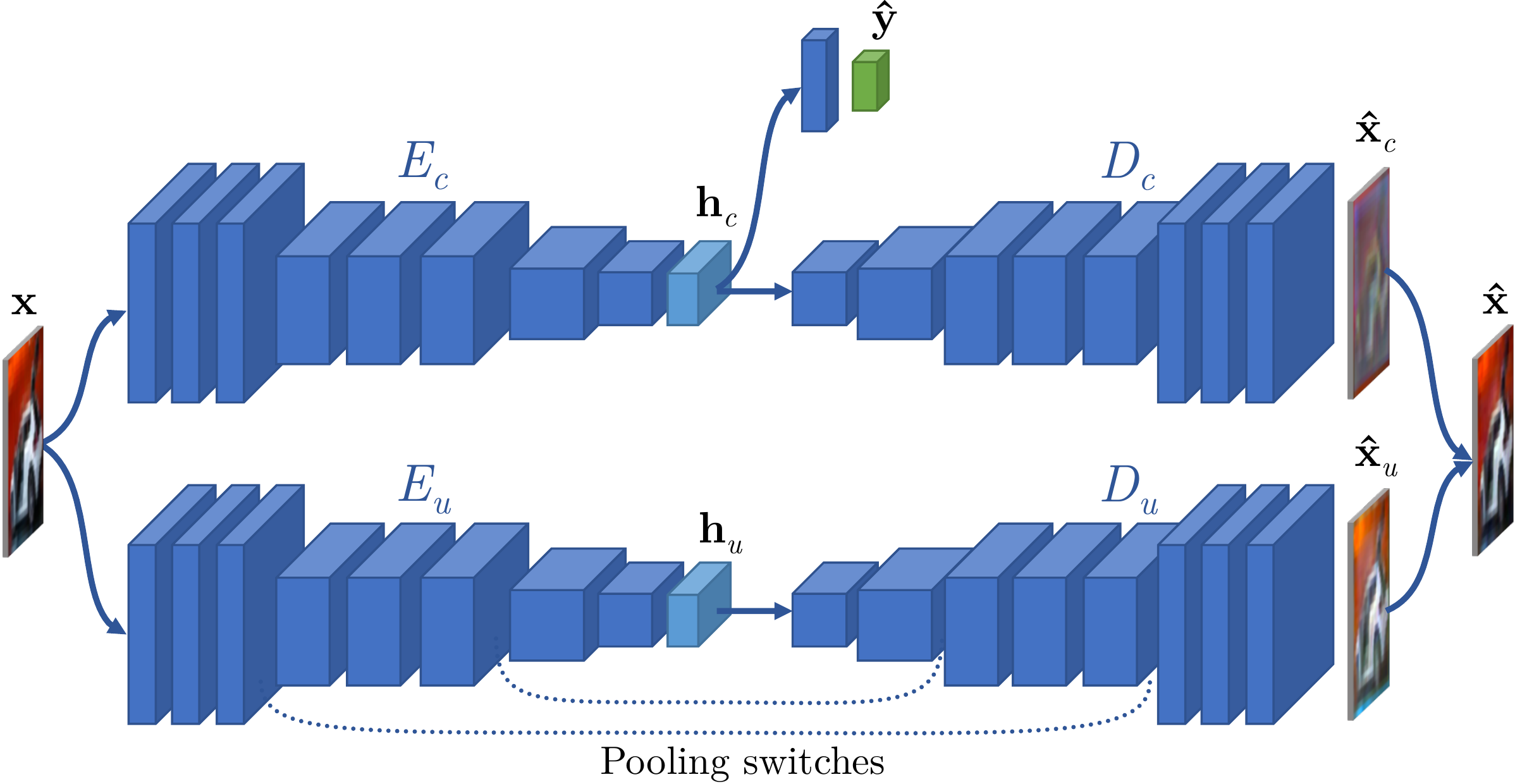}
    \caption{Example of HybridNet architecture where an original classifier (ConvLarge) constitutes $E_c$ and has been mirrored to create $D_c$ and duplicated for $E_u$ and $D_u$, with the addition of unpooling in the discriminative branch}
    \label{fig:cifar10-archi}
\end{figure}

\section{Experiments}
\label{sec:experiements}

In this section, we will study and validate the behavior of our novel framework. We first perform ablation studies to validate the architecture and loss terms of the model.  We also propose visualizations of the behavior of the model in various configurations, before demonstrating the capability of HybridNet to obtain state-of-the-art results.

In these experiments, we use three image datasets: SVHN~\cite{netzer2011reading}, CIFAR-10~\cite{cifar10} and STL-10~\cite{Coates2011}.
Both SVHN and CIFAR-10 are 10-classes datasets of $32\times 32$ pixels images. SVHN has 73,257 images for training, 26,032 for testing and 531,131 extra images used only as unlabeled data. CIFAR-10 has 50,000 training images and 10,000 testing images. For our semi-supervised experiments, we only keep $N$ labeled training samples (with $\nicefrac{N}{10}$ samples per class) while the rest of the data is kept unlabeled, as is commonly done.
STL-10 have the same 10 classes as CIFAR-10 with
$96\times 96$ pixels images. It is designed for semi-supervised learning since it contains 10 folds of 1,000 labeled training images, 100,000 unlabeled training images
and 8,000 test images with labels.

\subsection{HybridNet framework validation}
\label{sec:as}

First, we propose a thorough analysis of the behavior of our model at two different levels: first by comparing it to baselines that we obtain when disabling parts of the architecture, and second by analyzing the contribution of the different terms of the training loss of HybridNet both quantitatively and through visualizations.

This study was mainly performed using the ConvLarge architecture~\cite{Rasmus2015} on CIFAR-10 since it's a very common setup used in recent semi-supervised experiments~\cite{Sajjadi2016,Laine2016,Tarvainen2017}. The design of the HybridNet version of this architecture follows Section~\ref{sec:architecture} (illustrated in Fig.~\ref{fig:cifar10-archi}) and uses Temporal Ensembling to produce stability targets $\vz$. Additional results are provided using an adapted version of ConvLarge for STL-10 with added blocks of convolutions and pooling.

Models are trained with Adam with a learning rate of 0.003 for 600 epochs with batches of 20 labeled images and 80 unlabeled ones. The various loss-weighting terms $\lambda$ of the general loss (Eq. (\ref{eq:full-loss})) could have been optimized on a validation set but for these experiments they were simply set so that the different loss terms have values of the same order of magnitude.
Thus, all $\lambda$ were set to either 0 or 1 if activated or not, except $\lambda_s$ set to 0 or 100.
All the details of the experiments --\,exact architecture, hyperparameters, optimization, \textit{etc.}\,-- are provided in the appendix.

\subsubsection{Ablation study of the architecture.}

We start this analysis by validating our architecture with an ablation study on CIFAR-10 with various number of labeled samples. By disabling parts of the model and training terms, we compare HybridNet to different baselines and validate the importance of combining both contributions of the paper: the architecture and the training method.

Results are presented in Table~\ref{table:ablation}. The classification and auto-encoder results are obtained with the same code and hyperparameters by simply disabling different losses and parts of the model: the classifier only use $E_c$ and $C$; and the auto-encoder (similar to~\cite{Zhao2016a}) only $E_c$, $D_c$ and $C$. For both, we can add the stability loss. The HybridNet architecture only uses the classification and reconstructions loss terms while the second result uses the full training loss.

\begin{table}[tb]
  \setlength{\tabcolsep}{4pt}
  \caption{Ablation study performed on CIFAR-10 with ConvLarge architecture}
  \label{table:ablation}
  \centering
  \begin{tabular}{llll}
    \toprule
                                              & \multicolumn{3}{c}{Labeled samples}           \\ \cmidrule{2-4}
Model                                         & 1000          & 2000          & 4000          \\ \midrule
Classification                                & 63.4          & 71.5          & 79.0          \\
Classification and stability                  & 65.6          & 74.6          & 81.3          \\ \cmidrule{1-4}
Auto-encoder                                & 65.0          & 73.6          & 79.8          \\
Auto-encoder and stability                  & 71.8          & 80.4          & 84.9          \\ \cmidrule{1-4}

HybridNet architecture                        & 63.2          & 74.0          & 80.3          \\
HybridNet architecture and full training loss & \textbf{74.1} & \textbf{81.6} & \textbf{86.6} \\ \bottomrule
  \end{tabular}
\end{table}

First, we can see that the HybridNet architecture alone already yields an improvement over the baseline and the auto-encoder, except at 1000 labels. This could be explained by the fact that with very few labels, the model fails to correctly separate the information between the two branches because of the faint classification signal, and the additional loss terms that control the training of HybridNet are even more necessary. Overall, the architecture alone does not provide an important gain since it is not guided to efficiently take advantage of the two branches, indeed, we see that the addition of the complete HybridNet loss allows the model to provide much stronger results, with an improvement of 6-7\,pts over the architecture alone, around 5-6\,pts better than the stability or auto-encoding baseline, and 7-10\,pts more than the supervised baseline. The most challenging baseline is the stabilized auto-encoder that manages to take advantage of the stability loss but from which we still improve by 1.2-2.8\,pts.

This ablation study demonstrates the capability of the HybridNet framework to surpass the different architectural baselines, and shows the importance of the complementarity between the two-branch architecture and the complete training loss.

\subsubsection{Importance of the various loss terms.} We now propose a more fine-grain study to look at the importance of each loss term of the HybridNet training described in Section~\ref{sec:training}, both through classification results and visualizations.

First, in Table~\ref{table:as-scores} we show classification accuracy on CIFAR-10 with 2000 labels and STL-10 with 1000 labels for numerous combinations of loss terms. These results demonstrates that each loss term has it's importance and that all of them cooperate in order to reach the final best result of the full HybridNet model. In particular the stability loss is an important element of the training but is not sufficient as shown by lines \textit{b} and \textit{f-h}, while the other terms bring an equivalent gain as shown by lines \textit{c-e}. Both those $\sim$5\,pts gains can be combined to work in concert and reach the final score line \textit{i} of a $\sim$10\,pts gain.

Second, to interpret how the branches behave we propose to visualizing the different reconstructions $\vxh_c$, $\vxh_u$ and $\vxh$ for different combinations of loss terms in Table~\ref{table:asviz}. With only the final reconstruction term (lines \textit{c}), the discriminative branch does not contribute to the reconstruction and is thus barely regularized by the reconstruction loss, showing little gain over the classification baseline. The addition of the intermediate reconstruction terms helps the discriminative branch to produce a weak reconstruction (lines \textit{d}) and is complemented by the branch balancing technique (lines \textit{e}) to produce balanced reconstructions in both branch. The stability loss (lines \textit{i}) adds little visual impact on $\vxh_c$, it has probably more impact on the quality of the latent representation $\vh_c$ and seems to help in making the discriminative features and classifier more robust with a large improvement of the accuracy.

\newcolumntype{R}[2]{%
    >{\adjustbox{angle=#1,lap=\width-(#2)}\bgroup}%
    l%
    <{\egroup}%
}

\colorlet{verylightgray}{gray!20}
\newcommand{\rowmidlinewc}{\arrayrulecolor{verylightgray}\cmidrule{1-8}
            \arrayrulecolor{black}}
\newcommand{\rowmidlinewcb}{\arrayrulecolor{verylightgray}\cmidrule{1-5}
            \arrayrulecolor{black}}

\newcommand{\rot}{\multicolumn{1}{R{65}{1em}}}
\newcommand{\n}{\color{RoyalBlue}}
\newcommand*\OK{\ding{51}}

\newcommand{\spa}{\ \;}
\newcommand{\spb}{\quad\;}

\captionsetup[table]{skip=0pt}
\begin{table}[tb]
  \caption{Detailed ablation studies when activating different terms and techniques of the HybridNet learning. These results are obtained with ConvLarge on CIFAR-10 with 2000 labeled samples and ConvLarge-like on STL-10 with 1000 labeled samples}
  \label{table:abl}
  \begin{subfigure}[t]{0.32\textwidth}
  \caption{Test accuracy (\%)}
  \label{table:as-scores}
  \centering
  \renewcommand{\arraystretch}{1.1}
    \begin{tabular}{@{}clllllcc}
    \toprule
    &\rot{$\mathcal L_\textrm{classif}$} & \rot{$\Omega_\textrm{stability}$} & \rot{$\mathcal L_\textrm{rec}$ {\scriptsize(hybrid)}} & \rot{$\mathcal L_\textrm{rec-inter}$} & \rot{$\mathcal L_\textrm{rec-balanced}$} & \rot{CIFAR-10} & \rot{STL-10} \\ \midrule
    \scriptsize \textit{a} & \OK &     &     &     &    & 71.5 & 65.6         \\
    \scriptsize \textit{b} &\OK & \OK &     &     &    & 74.6 & 69.8         \\
    \rowmidlinewc
    \scriptsize \textit{c} &\OK &     & \OK &     &    & 72.4 & 67.8         \\
    \scriptsize \textit{d} &\OK &     & \OK & \OK &    & 74.0 & --         \\
    \scriptsize \textit{e} &\OK &     & \OK & \OK & \OK & 75.2 & --         \\
    \rowmidlinewc
    \scriptsize \textit{f} &\OK & \OK & \OK &     &    & 77.7 & 71.5           \\
    \scriptsize \textit{g} &\OK & \OK & \OK &     & \OK & 77.4 & --         \\
    \scriptsize \textit{h} &\OK & \OK & \OK & \OK &    & 80.8 & 72.2          \\
    \scriptsize \textit{i} &\OK & \OK & \OK & \OK & \OK & \textbf{81.6} & \textbf{74.1} \\
     \bottomrule
  \end{tabular}
  \end{subfigure}
  \hfill
  \begin{subfigure}[t]{0.675\textwidth}
  \caption{Visualization of partial and combined reconstructions}
  \label{table:asviz}
	\centering

  \setlength{\tabcolsep}{1.1pt}
  \renewcommand{\arraystretch}{1.28}
    \begin{tabular}{@{}cllllc@{}}
    \toprule
    &\rot{$\mathcal L_\textrm{rec}$ {\scriptsize(hybrid)}} & \rot{$\mathcal L_\textrm{rec-inter}$} & \rot{$\mathcal L_\textrm{rec-balanced}$} & \rot{$\Omega_\textrm{stability}$} &
    $\vx \spa \vxh_c \spa \vxh_u \spa \vxh \spb \vx \spa \vxh_c \spa \vxh_u \spa \vxh \spb \vx \spa \vxh_c \spa \vxh_u \spa \vxh $ \\ \midrule
    \scriptsize \textit{c} &\OK &     &     &     & \multirow{4}{*}{\includegraphics[width=6.2cm]{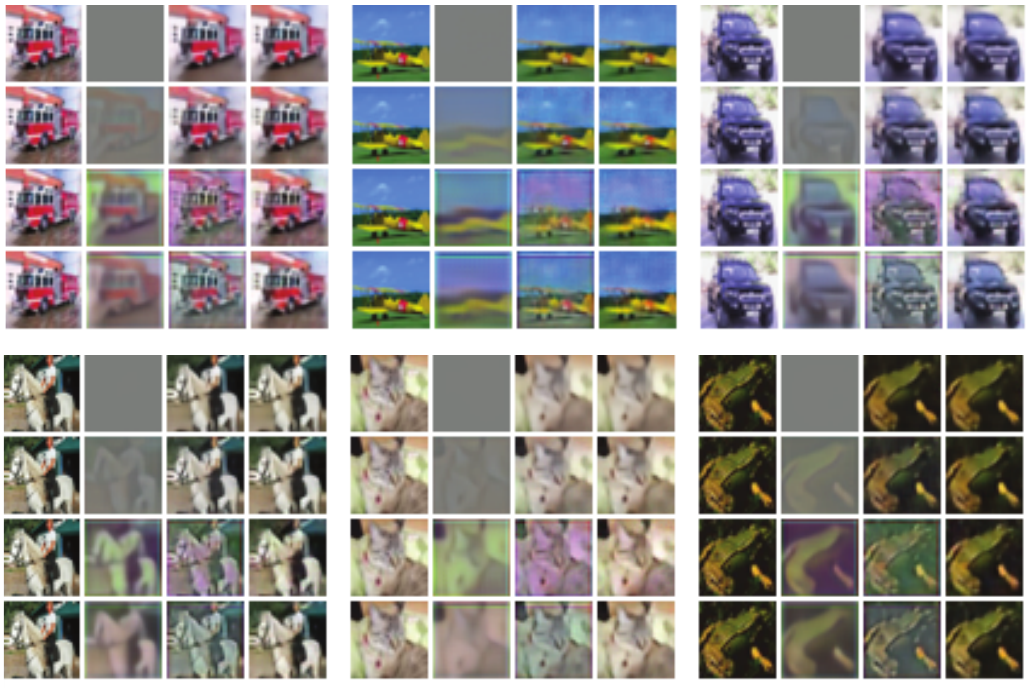}\!}          \\
    \scriptsize \textit{d} &\OK & \OK    &     &     &           \\
    \scriptsize \textit{e} &\OK & \OK    & \OK    &     &           \\
    \scriptsize \textit{i} &\OK & \OK    & \OK    & \OK    &           \\
    \rowmidlinewcb
    \scriptsize \textit{c} & \OK &        &     &     &           \\
    \scriptsize \textit{d} &\OK & \OK    &     &     &           \\
    \scriptsize \textit{e} &\OK & \OK    & \OK    &     &           \\
    \scriptsize \textit{i} &\OK & \OK    & \OK    & \OK    &           \\
     \bottomrule
  \end{tabular}

	\end{subfigure}
\end{table}
\captionsetup[table]{skip=10pt}

\subsubsection{Visualization of information separation on CIFAR-10 and STL-10.} 
\begin{figure}[tb]
	\centering
	\includegraphics[width=\textwidth]{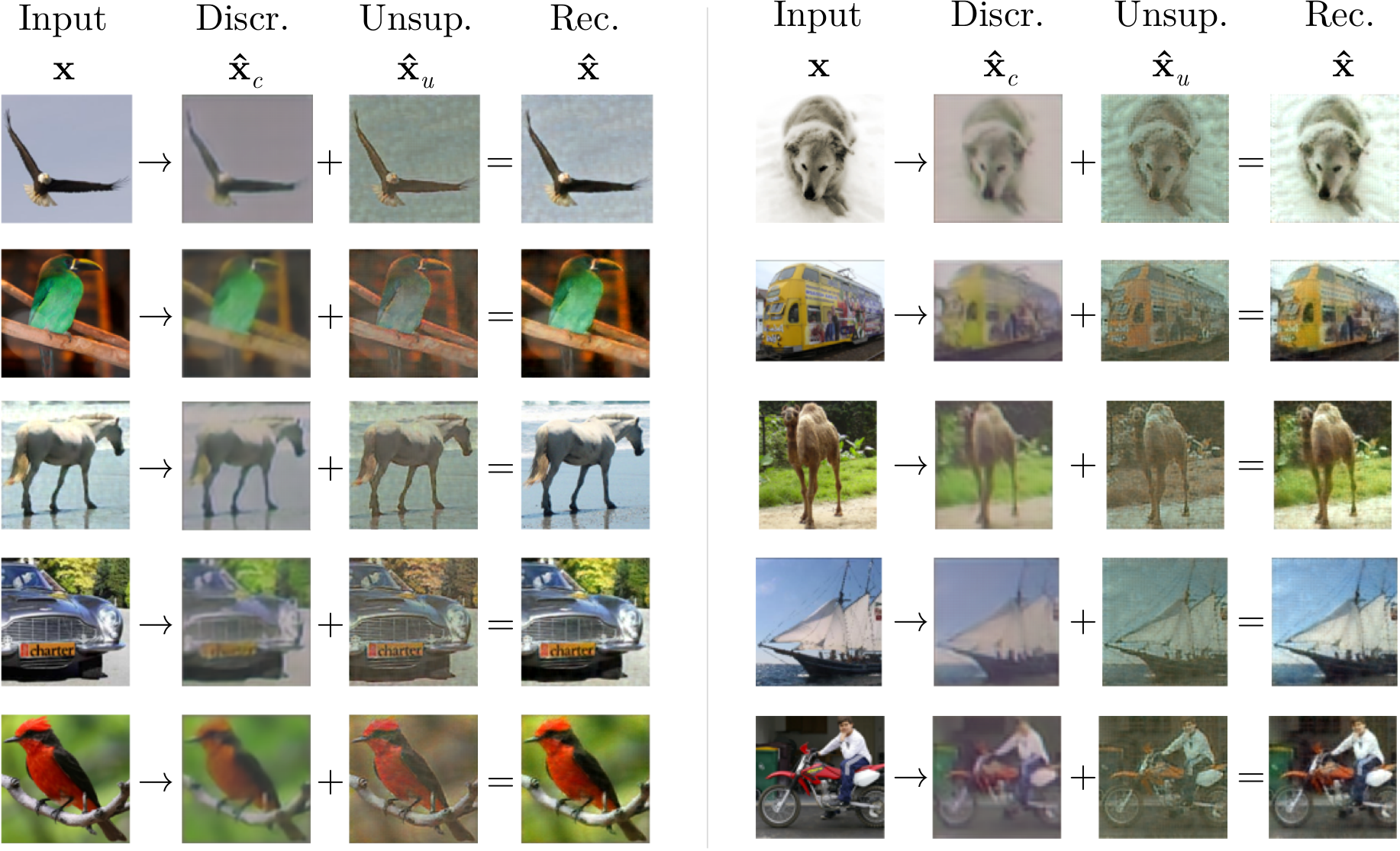}
    \caption{Visualizations of input, partial and final reconstructions of STL-10 images using a HybridNet model derived from a ConvLarge-like architecture}
    \label{fig:viz-stl}
\end{figure}

Overall, we can see in Table~\ref{table:asviz} lines \textit{i} that thanks to the full HybridNet training loss, the information is correctly separated between $\vxh_c$ and $\vxh_u$ than both contribute somewhat equally while specializing on different type of information. For example, for the blue car, $\vxh_c$ produces a blurry car with approximate colors, while $\vxh_u$ provides both shape details and exact color information. For nicer visualizations, we also show reconstructions of the full HybridNet model trained on STL-10 which has larger images in Fig.~\ref{fig:viz-stl}. These confirm the observations on CIFAR-10 with a  very good final reconstruction composed of a rough reconstruction that lacks texture and color details from the discriminative branch, completed by low-level details of shape, texture, writings, color correction and background information from the unsupervised branch.

\subsection{State-of-the-art comparison}
\label{sec:cifar10sota}

After studying the behavior of this novel architecture, we propose to demonstrate its effectiveness and capability to produce state-of-the-art results for semi-supervised learning on three datasets: SVHN, CIFAR-10 and STL-10.

We use ResNet architectures to constitute the supervised encoder $E_c$ and classifier $C$; and augment them with a mirror decoder $D_c$ and an unsupervised second branch containing an encoder $E_u$ and a decoder $D_u$ using the same architecture. For SVHN and CIFAR-10, we use the small ResNet from~\cite{Gastaldi2017}, which is used in Mean Teacher~\cite{Tarvainen2017} and currently achieves state-of-the-art results on CIFAR-10. For STL-10, we upscale the images to 224$\times$224\,px and use a regular ResNet-50 pretrained on the Places dataset.

We trained HybridNet with the training method described in Section~\ref{sec:training}, including Mean Teacher to produce stability targets $\vz\kk$. The training protocol follow exactly the protocol of Mean Teacher~\cite{Tarvainen2017} for CIFAR-10 and a similar one for SVHN and STL-10 for which \cite{Tarvainen2017} does not report results with ResNet. The hyperparameters added in HybridNet, \textit{i.e.} the weights of the reconstruction terms (final and intermediate), were coarsely adjusted on a validation set (we tried values 0.25, 0.5 and 1.0 for both of them). Details are in the appendix.

The results of these experiments are presented in Table~\ref{table:cifar10}.
We can see the huge performance boost obtained by HybridNet compared to the ResNet baselines, in particular with CIFAR-10 with 1000 labels where the error rate goes from 45.2\% to 8.81\%, which demonstrates the large benefit of our regularizer. HybridNet also improves over the strong Mean Teacher baseline~\cite{Tarvainen2017}, with an improvement of 1.29\,pt with 1000 labeled samples on CIFAR-10, and 0.9\,pt on STL-10. We also significantly improve over other stability-based approaches~\cite{Sajjadi2016,Laine2016}, and over the Ladder Networks~\cite{Rasmus2015} and GAN-based techniques~\cite{Springenberg2015,Salimans2016}.

These results demonstrate the capability of HybridNet to apply to large residual architecture --\,that are very common nowadays\,-- and to improve over baselines that already provided very good performance.

\begin{table}[tb]
  \setlength{\tabcolsep}{4pt}
  \caption{Results on CIFAR-10, STL-10 and SVHN using a ResNet-based HybridNet. ``Mean Teacher ResNet'' is our classification \& stability baseline; results marked with ${}^*$ are not reported in the original paper and were obtained ourselves 
  }
  \label{table:cifar10}
  \centering
  \begin{tabular}{@{}lrrrrrr@{}}
    \toprule
    Dataset                                    & \multicolumn{3}{c}{CIFAR-10} & \multicolumn{2}{c}{SVHN} & $\!\!\!$STL-10 \\
    \cmidrule{2-7}
    Nb. labeled images & 1000  & 2000 & 4000 & 500 & 1000 & \multicolumn{1}{c}{1000} \\
    \midrule
    SWWAE~\cite{Zhao2016a}                         &   &   &   & & 23.56 & 25.67 \\
    Ladder Network~\cite{Rasmus2015}                  &&       & 20.40      \\
    Improved GAN~\cite{Salimans2016}              & 21.83 & 19.61 & 18.63 & 18.44 & 8.11 &  \\
    CatGAN~\cite{Springenberg2015}                    &&       & 19.58      \\
    Stability regularization~\cite{Sajjadi2016}    &&       & 11.29  & 6.03 & &      \\
    Temporal Ensembling~\cite{Laine2016}           &&       & 12.16 & 5.12 & 4.42  \\
    Mean Teacher ConvLarge~\cite{Tarvainen2017}              & 21.55 & 15.73 & 12.31 & 4.18 & 3.95 \\
    Mean Teacher ResNet~\cite{Tarvainen2017}          & 10.10  &      & 6.28 & ${}^*$2.33 & ${}^*$2.05 & ${}^*$16.8        \\
    \cmidrule{1-7}
    ResNet baseline~\cite{Gastaldi2017}						 & 45.2 & 24.3 & 15.45 & 12.27 & 9.56 & 18.0  \\
    \textbf{HybridNet [ours]}  & \textbf{8.81} & \textbf{7.87} & \textbf{6.09} & \textbf{1.85} & \textbf{1.80} & \textbf{15.9} \\
    \bottomrule
  \end{tabular}
\end{table}

\section{Conclusion}

In this paper, we described a novel semi-supervised framework called HybridNet that proposes an auto-encoder-based architecture with two distinct paths that  separate the discriminative information useful for classification from the remaining information that is only useful for reconstruction. This architecture is accompanied by a loss and training technique that allows the architecture to behave in the desired way. In the experiments, we validate the significant performance boost brought by HybridNet in comparison with several other common architectures that use reconstruction losses and stability. We also show that HybridNet is able to produce state-of-the-art results on multiple datasets.

With two latent representations that explicitly encode classification information on one side and the remaining information on the other side, our model may be seen as a competitor to the fully reversible RevNets models recently proposed, that implicitly encode both types of information.
We plan to further explore the relationships between these approaches.

\textbf{Acknowledgements.} This work was funded by grant DeepVision (ANR-15-CE23-0029-02, STPGP-479356-15), a joint French/Canadian call by ANR \& NSERC.

\bibliographystyle{splncs}
\bibliography{biblio}

\begin{thebibliography}{10}

\bibitem{krizhevsky2012imagenet}
Krizhevsky, A., Sutskever, I., Hinton, G.E.:
\newblock Imagenet classification with deep convolutional neural networks.
\newblock In: Advances in Neural Information Processing Systems (NIPS). (2012)

\bibitem{he2016deep}
He, K., Zhang, X., Ren, S., Sun, J.:
\newblock Deep residual learning for image recognition.
\newblock In: IEEE Conference on Computer Vision and Pattern Recognition
  (CVPR). (2016)

\bibitem{Durand_WILDCAT_CVPR_2017}
Durand, T., Mordan, T., Thome, N., Cord, M.:
\newblock {WILDCAT: Weakly Supervised Learning of Deep ConvNets for Image
  Classification, Pointwise Localization and Segmentation}.
\newblock In: IEEE Conference on Computer Vision and Pattern Recognition
  (CVPR). (2017)

\bibitem{dai2016r}
Dai, J., Li, Y., He, K., Sun, J.:
\newblock R-fcn: Object detection via region-based fully convolutional
  networks.
\newblock In: Advances in Neural Information Processing Systems (NIPS). (2016)

\bibitem{redmon2016you}
Redmon, J., Divvala, S., Girshick, R., Farhadi, A.:
\newblock You only look once: Unified, real-time object detection.
\newblock In: IEEE Conference on Computer Vision and Pattern Recognition
  (CVPR). (2016)

\bibitem{Mordan2017}
Mordan, T., Thome, N., Cord, M., Henaff, G.:
\newblock {Deformable Part-based Fully Convolutional Network for Object
  Detection}.
\newblock In: {British Machine Vision Conference (BMVC)}. (2017)

\bibitem{deeplab}
Chen, L.C., Papandreou, G., Kokkinos, I., Murphy, K., Yuille, A.L.:
\newblock Deeplab: Semantic image segmentation with deep convolutional nets,
  atrous convolution, and fully connected crfs.
\newblock {IEEE} Transactions on Pattern Analysis and Machine Intelligence
  (TPAMI) (2018)

\bibitem{Martin2018}
Engilberge, M., Chevallier, L., P{\'{e}}rez, P., Cord, M.:
\newblock Finding beans in burgers: Deep semantic-visual embedding with
  localization.
\newblock IEEE Conference on Computer Vision and Pattern Recognition (CVPR)
  (2018)

\bibitem{Carvalho2018}
Carvalho, M., Cad{\`{e}}ne, R., Picard, D., Soulier, L., Thome, N., Cord, M.:
\newblock Cross-modal retrieval in the cooking context: Learning semantic
  text-image embeddings.
\newblock Special Interest Group on Information Retrieval (SIGIR) (2018)

\bibitem{benyounescadene2017mutan}
Ben-Younes, H., Cad{\`{e}}ne, R., Thome, N., Cord, M.:
\newblock Mutan: Multimodal tucker fusion for visual question answering.
\newblock IEEE International Conference on Computer Vision (ICCV) (2017)

\bibitem{weightdecay}
Krogh, A., Hertz, J.A.:
\newblock A simple weight decay can improve generalization.
\newblock In: Advances in Neural Information Processing Systems (NIPS). (1992)

\bibitem{batchnorm}
Ioffe, S., Szegedy, C.:
\newblock Batch normalization: Accelerating deep network training by reducing
  internal covariate shift.
\newblock Journal of Machine Learning Research (JMLR) (2016)

\bibitem{dropout}
Srivastava, N., Hinton, G., Krizhevsky, A., Sutskever, I., Salakhutdinov, R.:
\newblock Dropout: A simple way to prevent neural networks from overfitting.
\newblock Journal of Machine Learning Research (JMLR) (2014)

\bibitem{Blot2018}
Blot, M., Robert, T., Thome, N., Cord, M.:
\newblock Shade: Information-based regularization for deep learning.
\newblock In: IEEE International Conference on Image Processing (ICIP). (2018)

\bibitem{bengio2007greedy}
Bengio, Y., Lamblin, P., Popovici, D., Larochelle, H.:
\newblock Greedy layer-wise training of deep networks.
\newblock In: Advances in Neural Information Processing Systems (NIPS). (2007)

\bibitem{hinton2006reducing}
Hinton, G.E., Salakhutdinov, R.R.:
\newblock Reducing the dimensionality of data with neural networks.
\newblock Science (2006)

\bibitem{Zhao2016a}
Zhao, J., Mathieu, M., Goroshin, R., LeCun, Y.:
\newblock {Stacked What-Where Auto-encoders}.
\newblock In: International Conference on Learning Representations Workshop
  (ICLR-W). (2016)

\bibitem{Zhang2016a}
Zhang, Y., Lee, K., Lee, H.:
\newblock Augmenting supervised neural networks with unsupervised objectives
  for large-scale image classification.
\newblock In: International Conference on Machine Learning (ICML). (2016)

\bibitem{Rasmus2015}
Rasmus, A., Berglund, M., Honkala, M., Valpola, H., Raiko, T.:
\newblock Semi-supervised learning with ladder networks.
\newblock In: Advances in Neural Information Processing Systems (NIPS). (2015)

\bibitem{Mallat2011}
Mallat, S.:
\newblock Group invariant scattering.
\newblock Communications on Pure and Applied Mathematics (CPAM) (2012)

\bibitem{Bruna:2013:ISC:2498740.2498892}
Bruna, J., Mallat, S.:
\newblock Invariant scattering convolution networks.
\newblock {IEEE} Transactions on Pattern Analysis and Machine Intelligence
  (TPAMI) (2013)

\bibitem{BiettiNIPS17}
Bietti, A., Mairal, J.:
\newblock {Group Invariance, Stability to Deformations, and Complexity of Deep
  Convolutional Representations}.
\newblock In: Advances in Neural Information Processing Systems (NIPS). (2017)

\bibitem{Sajjadi2016}
Sajjadi, M., Javanmardi, M., Tasdizen, T.:
\newblock Regularization with stochastic transformations and perturbations for
  deep semi-supervised learning.
\newblock In: Advances in Neural Information Processing Systems (NIPS). (2016)

\bibitem{Laine2016}
Laine, S., Aila, T.:
\newblock Temporal ensembling for semi-supervised learning.
\newblock In: International Conference on Learning Representations (ICLR).
  (2017)

\bibitem{Tarvainen2017}
Tarvainen, A., Valpola, H.:
\newblock Mean teachers are better role models: Weight-averaged consistency
  targets improve semi-supervised deep learning results.
\newblock In: Advances in Neural Information Processing Systems (NIPS). (2017)

\bibitem{Zhu2005}
Zhu, X.:
\newblock Semi-supervised learning literature survey.
\newblock Technical Report 1530, Computer Sciences, University of
  Wisconsin-Madison (2005)

\bibitem{Ranzato2008}
Ranzato, M., Szummer, M.:
\newblock Semi-supervised learning of compact document representations with
  deep networks.
\newblock In: International Conference on Machine Learning (ICML). (2008)

\bibitem{Ranzato2007b}
Ranzato, M., Huang, F.J., Boureau, Y.L., LeCun, Y.:
\newblock Unsupervised learning of invariant feature hierarchies with
  applications to object recognition.
\newblock In: IEEE Conference on Computer Vision and Pattern Recognition
  (CVPR). (June 2007)

\bibitem{vincent2008extracting}
Vincent, P., Larochelle, H., Bengio, Y., Manzagol, P.A.:
\newblock Extracting and composing robust features with denoising autoencoders.
\newblock In: International Conference on Machine Learning (ICML). (2008)

\bibitem{Ranzato2007}
Ranzato, M., Poultney, C., Chopra, S., Lecun, Y.:
\newblock Efficient learning of sparse representations with an energy-based
  model.
\newblock In: Advances in Neural Information Processing Systems (NIPS). (2007)

\bibitem{Larochelle2008}
Larochelle, H., Bengio, Y.:
\newblock Classification using discriminative restricted boltzmann machines.
\newblock In: International Conference on Machine Learning (ICML). (2008)

\bibitem{kingma2014semi}
Kingma, D.P., Mohamed, S., Rezende, D.J., Welling, M.:
\newblock Semi-supervised learning with deep generative models.
\newblock In: Advances in Neural Information Processing Systems (NIPS). (2014)

\bibitem{erhan2010does}
Erhan, D., Bengio, Y., Courville, A., Manzagol, P.A., Vincent, P., Bengio, S.:
\newblock Why does unsupervised pre-training help deep learning?
\newblock Journal of Machine Learning Research (JMLR) (2010)

\bibitem{Goh_NIPS13}
Goh, H., Thome, N., Cord, M., Lim, J.H.:
\newblock {Top-down regularization of deep belief networks}.
\newblock In: Advances in Neural Information Processing Systems (NIPS). (2013)

\bibitem{NIPS2014_5423}
Goodfellow, I., Pouget-Abadie, J., Mirza, M., Xu, B., Warde-Farley, D., Ozair,
  S., Courville, A., Bengio, Y.:
\newblock Generative adversarial nets.
\newblock In: Advances in Neural Information Processing Systems (NIPS). (2014)

\bibitem{Makhzani2016}
Makhzani, A., Shlens, J., Jaitly, N., Goodfellow, I.:
\newblock Adversarial autoencoders.
\newblock In: International Conference on Learning Representations (ICLR).
  (2016)

\bibitem{hastie_09_elements-of.statistical-learning}
Hastie, T., Tibshirani, R., Friedman, J.:
\newblock {The Elements of Statistical Learning}.
\newblock Springer (2009)

\bibitem{TheriaultCVPR13}
Th\'{e}riault, C., Thome, N., Cord, M.:
\newblock {D}ynamic {S}cene {C}lassification: {L}earning {M}otion {D}escriptors
  with {S}low {F}eatures {A}nalysis.
\newblock In: IEEE Conference on Computer Vision and Pattern Recognition
  (CVPR). (2013)

\bibitem{goodfellow2014explaining}
Goodfellow, I.J., Shlens, J., Szegedy, C.:
\newblock Explaining and harnessing adversarial examples.
\newblock In: International Conference on Learning Representations (ICLR).
  (2015)

\bibitem{miyato2015distributional}
Miyato, T., Maeda, S.i., Koyama, M., Nakae, K., Ishii, S.:
\newblock Distributional smoothing with virtual adversarial training.
\newblock In: International Conference on Learning Representations (ICLR).
  (2016)

\bibitem{Wojna2017}
Wojna, Z., Ferrari, V., Guadarrama, S., Silberman, N., Chen, L.C., Fathi, A.,
  Uijlings, J.:
\newblock The devil is in the decoder.
\newblock In: {British Machine Vision Conference (BMVC)}. (2017)

\bibitem{dumoulin2016guide}
Dumoulin, V., Visin, F.:
\newblock A guide to convolution arithmetic for deep learning.
\newblock Technical Report (2016)

\bibitem{NIPS2017_6816}
Gomez, A.N., Ren, M., Urtasun, R., Grosse, R.B.:
\newblock The reversible residual network: Backpropagation without storing
  activations.
\newblock In: Advances in Neural Information Processing Systems (NIPS). (2017)

\bibitem{jacobsen:hal-01712808}
Jacobsen, J.H., Smeulders, A., Oyallon, E.:
\newblock {i-RevNet: Deep Invertible Networks}.
\newblock In: International Conference on Learning Representations (ICLR).
  (2018)

\bibitem{swe:spie95}
Sweldens, W.:
\newblock The lifting scheme: A new philosophy in biorthogonal wavelet
  constructions.
\newblock In: Wavelet Applications in Signal and Image Processing III. (1995)

\bibitem{wavelets}
Mallat, S.G., Peyr\'{e}, G.:
\newblock {A wavelet tour of signal processing : the sparse way}.
\newblock Academic Press (2009)

\bibitem{zeiler2014visualizing}
Zeiler, M.D., Fergus, R.:
\newblock Visualizing and understanding convolutional networks.
\newblock In: European Conference on Computer Vision (ECCV). (2014)

\bibitem{netzer2011reading}
Netzer, Y., Wang, T., Coates, A., Bissacco, A., Wu, B., Ng, A.Y.:
\newblock Reading digits in natural images with unsupervised feature learning.
\newblock In: NIPS workshop on deep learning and unsupervised feature learning.
  (2011)

\bibitem{cifar10}
Krizhevsky, A., Hinton, G.:
\newblock Learning multiple layers of features from tiny images.
\newblock Technical Report (2009)

\bibitem{Coates2011}
Coates, A., Ng, A., Lee, H.:
\newblock An analysis of single-layer networks in unsupervised feature
  learning.
\newblock In: International Conference on Artificial Intelligence and
  Statistics (AISTATS). (2011)

\bibitem{Gastaldi2017}
Gastaldi, X.:
\newblock Shake-shake regularization of 3-branch residual networks.
\newblock In: International Conference on Learning Representations Workshop
  (ICLR-W). (2017)

\bibitem{Springenberg2015}
Springenberg, J.T.:
\newblock {Unsupervised and Semi-supervised Learning with Categorical
  Generative Adversarial Networks}.
\newblock In: International Conference on Learning Representations (ICLR).
  (2016)

\bibitem{Salimans2016}
Salimans, T., Goodfellow, I., Zaremba, W., Cheung, V., Radford, A., Chen, X.,
  Chen, X.:
\newblock Improved techniques for training gans.
\newblock In Lee, D.D., Sugiyama, M., Luxburg, U.V., Guyon, I., Garnett, R.,
  eds.: Advances in Neural Information Processing Systems (NIPS).
\newblock Curran Associates, Inc. (2016)

\end{thebibliography}

\newpage
\appendix

\section{Additional visualizations of HybridNet}

\begin{figure}
	\centering
    \includegraphics[width=\textwidth]{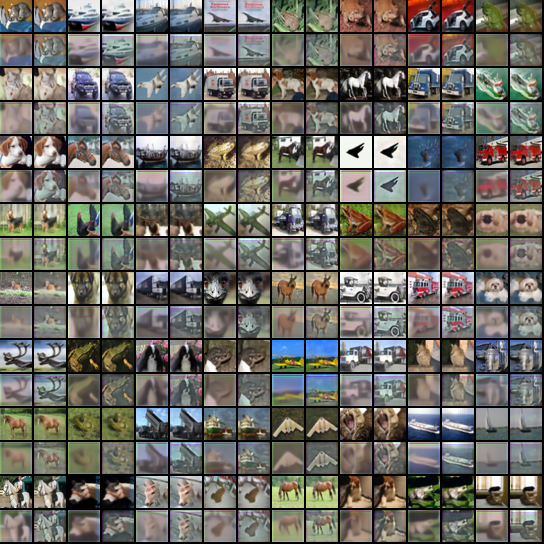}
    \caption[]{Example of visualizations for a ConvLarge-based HybridNet (complete training loss) on CIFAR-10. For each input image, there is block of 4 images on the figure with the following organization: $\begin{bmatrix}\vx&\vxh\\\vxh_c&\vxh_u\end{bmatrix}$}
    \label{fig:cifar10}
\end{figure}

\begin{figure}
	\centering
    \includegraphics[width=\textwidth]{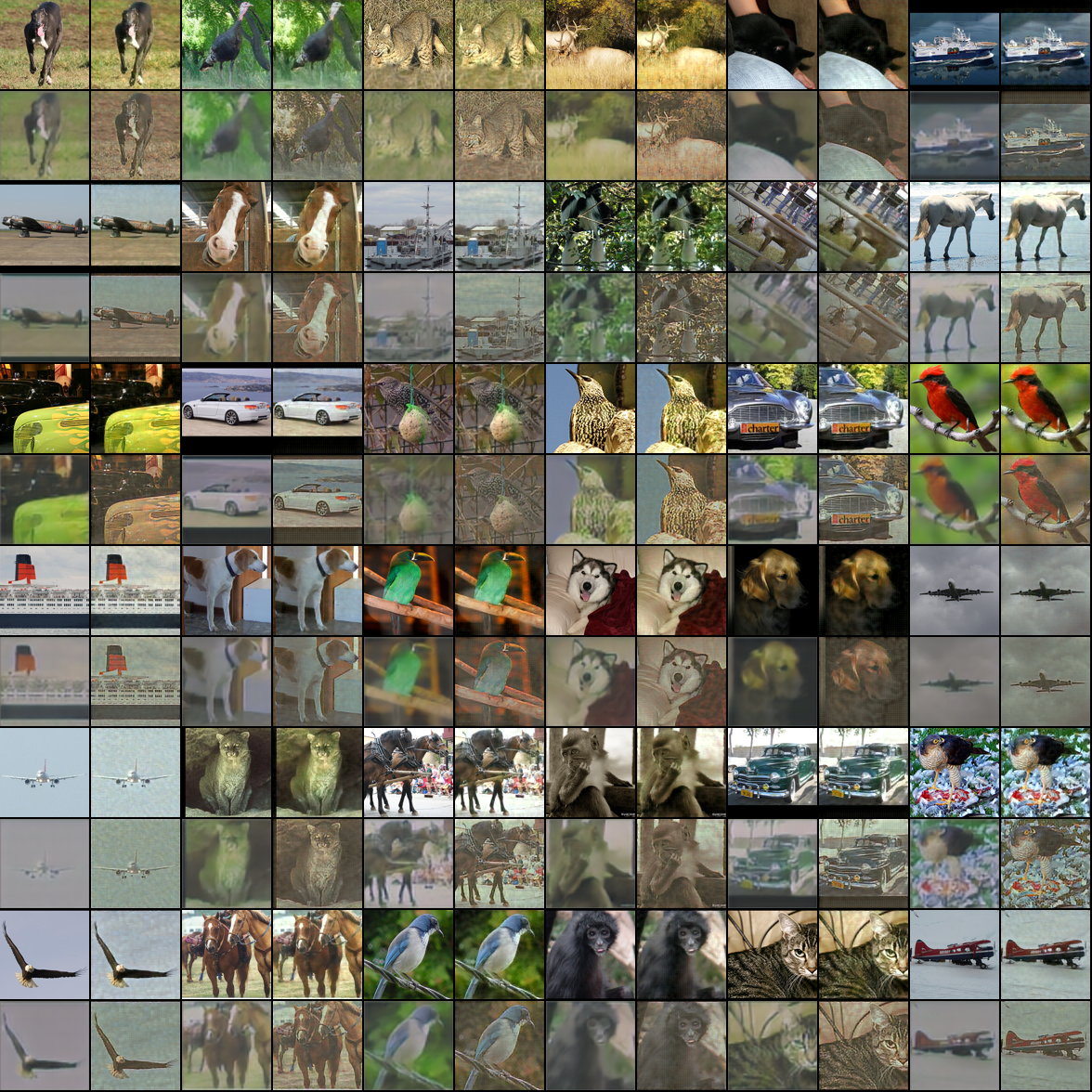}
    \caption[]{Example of visualizations for a ConvLarge-like-based HybridNet (complete training loss) on STL-10. For each input image, there is block of 4 images on the figure with the following organization: $\begin{bmatrix}\vx&\vxh\\\vxh_c&\vxh_u\end{bmatrix}$}
    \label{fig:stl10}
\end{figure}

Additional visualizations of HybridNet behavior are presented on Fig.~\ref{fig:cifar10} for CIFAR-10 and Fig.~\ref{fig:stl10} for STL-10.

\FloatBarrier

\section{Experimental setup for ConvLarge on CIFAR-10}
\subsection{Data preprocessing and model architecture}

Input images are data-augmented with a random translation of a maximum of 2 pixels with mirror padding to fill-in the missing pixels, and randomly flipped. This constitutes the input $\vx$. We also add a Gaussian noise on $\vx$ $(\sigma=0.15)$ to obtain $\tilde\vx$ that is fed into the model.
The model's architecture is described on Table~\ref{tab:convlarge}.

\subsection{Training details}

The training method is similar to the one presented in recent paper using Conv\-Large~\cite{Sajjadi2016,Laine2016,Tarvainen2017}.

The model is optimized with Adam during 60,000 batches (which corresponds to various number of epochs depending on the number of labeled images), with batches of 80 unlabeled samples and 20 labeled samples.

The weights of the various loss terms and the optimizer's parameters have base values and are varied over the training similarly to previous work using this model. The parameters' values and variations are summarized in Table~\ref{tab:convlargesched}. For the ablation study, parts of the model are removed and/or some weights are set to 0.

\section{Experimental setup for ConvLarge-like on STL-10}
\subsection{Model architecture}

Input images are data-augmented with a random translation of a maximum of 12 pixels with mirror padding to fill-in the missing pixels, and randomly flipped. This constitutes the input $\vx$. We also add a Gaussian noise on $\vx$ $(\sigma=0.15)$ to obtain $\tilde\vx$ that is fed into the model.
The model's architecture is detailed in Table~\ref{tab:convlargestl}.

\subsection{Training details}

The model is optimized with Adam during 150,000 batches (corresponding to 300 epochs over the labeled images, 48 epochs over the unlabeled images), with batches of 30 unlabeled samples and 2 labeled samples.
Hyperparameters' values and scheduling over training are detailed in Table~\ref{tab:convlargestlsched}.

\section{Experimental setup for ResNet on CIFAR-10 and SVHN}

The experimental setup described below follows the one described in~\cite{Tarvainen2017} for a fair comparison.

\subsection{Model architecture}

The data preprocessing simply consists in a classic per-color-channel mean-variance standardization. Images are data-augmented using random translation of a maximum of 4 pixels with mirror padding to fill-in the missing pixels and random flip. For SVHN, we disable image mirroring for obvious reasons.

\begin{figure}[!tb]
	\centering
    \includegraphics[width=0.45\textwidth]{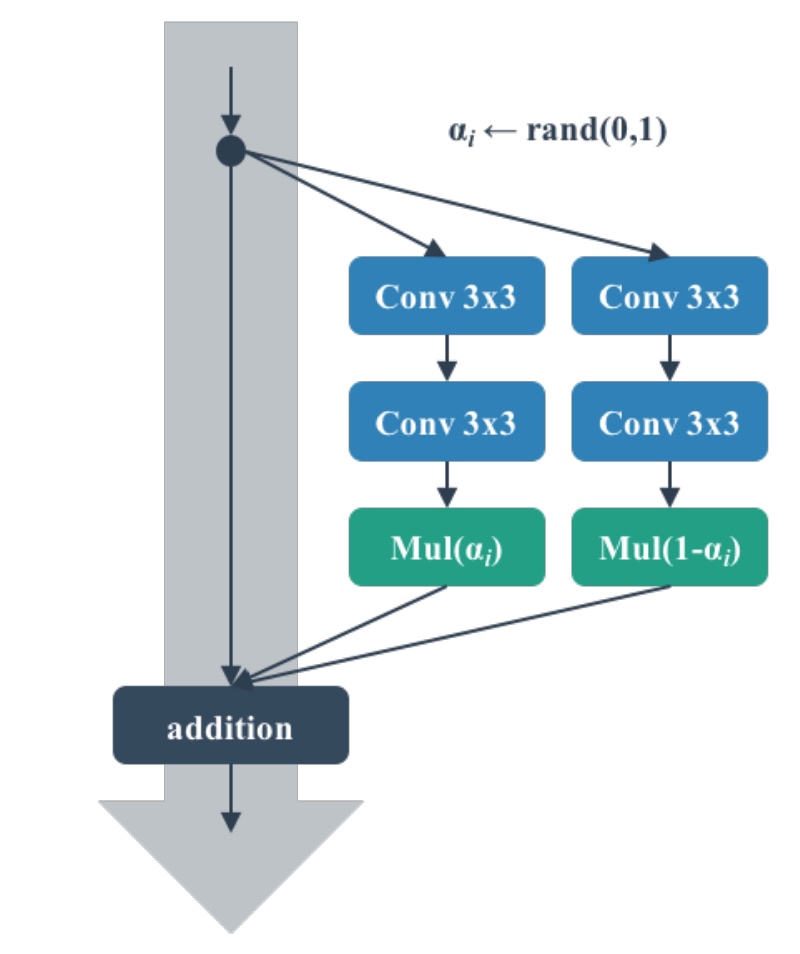}
    \caption[]{Shake-Shake building block}
    \label{fig:shake}
\end{figure}

We use the ResNet architecture with Shake-Shake building blocks described in~\cite{Gastaldi2017}. A Shake-Shake building block consists of 2~similar branches each containing 2~convolutions, with the first one possibly having a stride greater that~1. The two branches are averaged with a weight $\alpha$ (see Fig.~\ref{fig:shake} for illustration and the original paper for details) before being added to the result of a residual connection.

A ``layer'' is constituted of 4 blocks with possibly the first one having a stride in its first convolution and all convolutions having the same number of channels.

To reverse a layer, we apply the same strategy as before. This means that only the last transposed convolution of a decoding layer will have a smaller number of channels and a ``stride'' larger than~1 to reverse the first convolution of the corresponding layer in the encoder.

The architecture of the HybridNet based on this ResNet is described in Table~\ref{tab:resnet}.

\subsection{Training details for CIFAR-10}

The model's training is based on the settings of~\cite{Tarvainen2017}. It is trained with Nesterov SGD with a base learning rate of 0.04 with a momentum of 0.9 over 300 epochs (one epoch correspond to one pass over the unlabeled images) with batches of 61 unlabeled images and 19 labeled images. Hyperparameters values and scheduling over training are detailed in Table~\ref{tab:resnetsched}.

\subsection{Training details for SVHN}

The model is trained with Nesterov SGD with a base learning rate of 0.04 with a momentum of 0.9 over 150 epochs (one epoch correspond to one pass over the unlabeled images) with batches of 265 unlabeled images and 15 labeled images. Hyperparameters values and scheduling over training are detailed in Table~\ref{tab:resnetschedsvhn} for SVHN.

\section{Experimental setup for ResNet on STL-10}

\subsection{Model architecture}

The model is a ResNet-50 pretraind on the Places dataset available at \url{https://github.com/CSAILVision/places365}. We did not use a model trained on ImageNet since the images of STL-10 have been extracted from ImageNet.

The data preprocessing simply consists in a classic per-color-channel mean-variance standardization. Images are data-augmented using random translation of a maximum of 30 pixels with mirror padding to fill-in the missing pixels and random flip.

\subsection{Training details}

The model is trained with Nesterov SGD with a base learning rate of 0.01 with a momentum of 0.9 over 350 epochs (one epoch correspond to one pass over the unlabeled images) with batches of 11 unlabeled images and 5 labeled images. Hyperparameters values and scheduling over training are detailed in Table~\ref{tab:resnetschedstl}.

\begin{table}[htbp]
\centering
\caption{Architecture of the HybridNet ConvLarge architecture for CIFAR-10}
\label{tab:convlarge}
\begin{threeparttable}
\setlength{\tabcolsep}{4pt}
\begin{tabular}{ l l l}
\toprule
\multicolumn{3}{c}{\textbf{Encoders $E_c$ and $E_u$}} \\
\midrule
Input & $\tilde\vx$ & $32\times 32\times 3$ \\
Convolution & $128$ filters, $3\times3$, \textit{same} padding & $32\times 32\times 128$ \\
Convolution & $128$ filters, $3\times3$, \textit{same} padding & $32\times 32\times 128$ \\
Convolution & $128$ filters, $3\times3$, \textit{same} padding & $32\times 32\times 128$ \\
Pooling   & Maxpool $2\times2$ & $16\times 16\times 128$ \\
Dropout   & $p=0.5$  & $16\times 16\times 128$ \\
Convolution & $256$ filters, $3\times3$, \textit{same} padding  & $16\times 16\times 256$ \\
Convolution & $256$ filters, $3\times3$, \textit{same} padding  & $16\times 16\times 256$ \\
Convolution & $256$ filters, $3\times3$, \textit{same} padding  & $16\times 16\times 256$ \\
Pooling & Maxpool $2\times2$  & $8\times 8\times 256$ \\
Dropout & $p=0.5$  & $8\times 8\times 256$ \\
Convolution & $512$ filters, $3\times3$, \textit{valid} padding  & $6\times 6\times 512$ \\
Convolution & $256$ filters, $1\times1$, \textit{same} padding & $6\times 6\times 256$ \\
Convolution & $128$ filters, $1\times1$, \textit{same} padding & $6\times 6\times 128$ \\
Output & $\vh_c$ or $\vh_u$ & $6\times 6\times 128$ \\

\toprule
\multicolumn{3}{c}{\textbf{Classifier $C$}}\\
\midrule
Input & $\vh_c$& $6\times 6\times 128$ \\
Pooling & Global average pool & $1\times 1\times 128$ \\
Fully connected & with Softmax & $10$ \\
Output & $\vyh$ & 10 \\
\toprule
\multicolumn{3}{c}{\textbf{Decoders $D_c$ and $D_u$}}\\
\midrule
Input & $\vh_c$ or $\vh_u$ & $6\times 6\times 128$ \\
TConvolution & $256$ filters, $1\times1$, \textit{same} padding  & $6\times 6\times 256$ \\
TConvolution & $512$ filters, $1\times1$, \textit{same} padding  & $6\times 6\times 512$ \\
TConvolution & $256$ filters, $3\times3$, \textit{valid} padding  & $8\times 8\times 256$ \\
Upsampling   & $2\times2$ (unpooling in $D_u$)  & $16\times 16\times 256$ \\
TConvolution & $256$ filters, $3\times3$, \textit{same} padding & $16\times 16\times 256$ \\
TConvolution & $256$ filters, $3\times3$, \textit{same} padding & $16\times 16\times 256$ \\
TConvolution & $128$ filters, $3\times3$, \textit{same} padding & $16\times 16\times 128$ \\
Upsampling & $2\times2$ (unpooling in $D_u$) & $32\times 32\times 128$ \\
TConvolution & $128$ filters, $3\times3$, \textit{same} padding  & $32\times 32\times 128$ \\
TConvolution & $128$ filters, $3\times3$, \textit{same} padding  & $32\times 32\times 128$ \\
TConvolution & $3$ filters, $3\times3$, \textit{same} padding  & $32\times 32\times 3$ \\
Output & $\vxh_c$ or $\vxh_u$ & $32\times 32 \times 3$ \\
\bottomrule
\end{tabular}
\begin{tablenotes}
TConvolution stands for ``transposed convolution''.\\ Each Convolution or TConvolution is followed by a Batch Normalization layer and a LeakyRELU of parameter $\alpha = 0.1$
\end{tablenotes}
\end{threeparttable}
\end{table}

\begin{table}[htbp]
\centering
\caption{Evolution of weights for ConvLarge on CIFAR-10}
\label{tab:convlargesched}
\begin{threeparttable}
\setlength{\tabcolsep}{4pt}
\begin{tabular}{ l l l}
\toprule
 & Value & Scheduling \\
\midrule
$\eta$ & 0.003 & Linear decrease to 0 over the last \nicefrac{1}{3} of the training \\
$\beta_1$ & 0.9 & Exponential decrease to 0.5 over the last \nicefrac{1}{5} of the training \\
$\lambda_c$ & 1 & Exponential increase from 0 over 800 first batches \\
$\lambda_s$ & 100 & Exponential increase from 0 over first \nicefrac{1}{4} of the training \\
& & and exponential decrease to 0 over the last \nicefrac{1}{5} of the training \\
$\lambda_r$ & 1 & Exponential decrease over the last 5\% of the training \\
\bottomrule
\end{tabular}
\begin{tablenotes}
$\eta$ is the learning rate, $\beta_1$ the first momentum of Adam, $\lambda_c$ the classification weight, $\lambda_s$ the stability weight, $\lambda_r$ the reconstructions weights.\\
Exponential decrease follows the function $\exp(-5t^2)$ with $t\in[0,1]$ from the start to the end of the decreasing interval. When increasing, $t$ goes from 1 to 0.
\end{tablenotes}
\end{threeparttable}
\end{table}

\begin{table}[htbp]
\centering
\caption{Architecture of the HybridNet ConvLarge-like architecture for STL-10}
\label{tab:convlargestl}
\begin{threeparttable}
\setlength{\tabcolsep}{4pt}
\begin{tabular}{ l l l}
\toprule
\multicolumn{3}{c}{\textbf{Encoders $E_c$ and $E_u$}} \\
\midrule
Input & $\tilde\vx$ & $96\times 96\times 3$ \\
Convolution & $64$ filters, $3\times3$, \textit{same} padding & $96\times 96\times 64$ \\
Convolution & $64$ filters, $3\times3$, \textit{same} padding & $96\times 96\times 64$ \\
Pooling   & Maxpool $2\times2$ & $48\times 48\times 64$ \\

Convolution & $128$ filters, $3\times3$, \textit{same} padding & $48\times 48\times 128$ \\
Convolution & $128$ filters, $3\times3$, \textit{same} padding & $48\times 48\times 128$ \\
Pooling   & Maxpool $2\times2$ & $24\times 24\times 128$ \\

Convolution & $256$ filters, $3\times3$, \textit{same} padding  & $24\times 24\times 256$ \\
Convolution & $256$ filters, $3\times3$, \textit{same} padding  & $24\times 24\times 256$ \\
Pooling   & Maxpool $2\times2$ & $12\times 12\times 256$ \\

Convolution & $256$ filters, $3\times3$, \textit{same} padding  & $12\times 12\times 256$ \\
Pooling & Maxpool $2\times2$  & $6\times 6\times 256$ \\

Output & $\vh_c$ or $\vh_u$ & $6\times 6\times 256$ \\

\toprule
\multicolumn{3}{c}{\textbf{Classifier $C$}}\\
\midrule
Input & $\vh_c$& $6\times 6\times 256$ \\

Convolution & $512$ filters, $4\times4$, \textit{valid} padding  & $3\times 3\times 512$ \\
Dropout & $p=0.5$  & $3\times 3\times 512$ \\
Convolution & $512$ filters, $1\times1$, \textit{same} padding & $3\times 3\times 512$ \\
Dropout & $p=0.5$  & $3\times 3\times 512$ \\
Convolution & $10$ filters, $1\times1$, \textit{same} padding & $3\times 3\times 10$ \\

Pooling & Global average pool & $1\times1\times10$ \\
Softmax &   & $10$ \\
Output & $\vyh$ & 10 \\
\toprule
\multicolumn{3}{c}{\textbf{Decoders $D_c$ and $D_u$}}\\
\midrule
Input & $\vh_c$ or $\vh_u$ & $6\times 6\times 256$ \\
Upsampling   & $2\times2$ (unpooling in $D_u$)  & $12\times 12\times 256$ \\
TConvolution & $256$ filters, $3\times3$, \textit{same} padding  & $12\times 12\times 256$ \\
Upsampling   & $2\times2$ (unpooling in $D_u$)  & $24\times 24\times 256$ \\
TConvolution & $256$ filters, $3\times3$, \textit{same} padding  & $24\times 24\times 256$ \\
TConvolution & $128$ filters, $3\times3$, \textit{same} padding  & $24\times 24\times 128$ \\
Upsampling   & $2\times2$ (unpooling in $D_u$)  & $48\times 48\times 128$ \\
TConvolution & $128$ filters, $3\times3$, \textit{same} padding  & $48\times 48\times 128$ \\
TConvolution & $64$ filters, $3\times3$, \textit{same} padding  & $48\times 48\times 64$ \\
Upsampling   & $2\times2$ (unpooling in $D_u$)  & $96\times 96\times 64$ \\
TConvolution & $64$ filters, $3\times3$, \textit{same} padding  & $96\times 96\times 64$ \\
TConvolution & $3$ filters, $3\times3$, \textit{same} padding  & $96\times 96\times 3$ \\
Output & $\vxh_c$ or $\vxh_u$ & $96\times 96 \times 3$ \\
\bottomrule
\end{tabular}
\begin{tablenotes}
TConvolution stands for ``transposed convolution''.\\ Each Convolution or TConvolution is followed by a Batch Normalization layer and a ELU activation.
\end{tablenotes}
\end{threeparttable}
\end{table}

\begin{table}[htbp]
\centering
\caption{Evolution of weights for ConvLarge-like on STL-10}
\label{tab:convlargestlsched}
\begin{threeparttable}
\setlength{\tabcolsep}{4pt}
\begin{tabular}{ l l l}
\toprule
 & Value & Scheduling \\
\midrule
$\eta$ & 0.001 & Linear decrease to 0 over the last \nicefrac{1}{10} of the training \\
$\beta_1$ & 0.9 & Constant \\
$\lambda_c$ & 1 & Exponential increase from 0 over 4000 first batches \\
$\lambda_s$ & 300 & Exponential increase from 0 over first \nicefrac{1}{4} of the training \\
& & and exponential decrease to 0 over the last \nicefrac{1}{4} of the training \\
$\lambda_r$ & 1 & Exponential decrease over the last 5\% of the training \\
\bottomrule
\end{tabular}
\begin{tablenotes}
$\eta$ is the learning rate, $\beta_1$ the first momentum of Adam, $\lambda_c$ the classification weight, $\lambda_s$ the stability weight, $\lambda_r$ the reconstructions weights.\\
Exponential decrease follows the function $\exp(-5t^2)$ with $t\in[0,1]$ from the start to the end of the decreasing interval. When increasing, $t$ goes from 1 to 0.
\end{tablenotes}
\end{threeparttable}
\end{table}

\begin{table}[htbp]
\centering
\caption{Architecture of the HybridNet ResNet architecture for CIFAR-10 and SVHN}
\label{tab:resnet}
\begin{threeparttable}
\setlength{\tabcolsep}{4pt}
\begin{tabular}{ l l l}
\toprule
\multicolumn{3}{c}{\textbf{Encoders $E_c$ and $E_u$}} \\
\midrule
Input & $\tilde\vx$ & $32\times 32\times 3$ \\
Convolution & $16$ filters, $3\times3$, \textit{same} padding & $32\times 32\times 16$ \\

Shake Shake layer & $4$ blocks, $96$ filters, $3\times3$,  stride 1 & $32\times 32\times 96$ \\
Shake Shake layer & $4$ blocks, $192$ filters, $3\times3$,  stride 2 & $16\times 16\times 192$ \\
Shake Shake layer & $4$ blocks, $384$ filters, $3\times3$,  stride 2 & $8\times 8\times 384$ \\
Output & $\vh_c$ or $\vh_u$ & $8\times 8\times 384$ \\

\toprule
\multicolumn{3}{c}{\textbf{Classifier $C$}}\\
\midrule
Input & $\vh_c$& $8\times 8\times 384$ \\
Pooling & Global average pool & $1\times 1\times 384$ \\
Fully connected & with Softmax & $10$ \\
Output & $\vyh$ & 10 \\
\toprule
\multicolumn{3}{c}{\textbf{Decoders $D_c$ and $D_u$}}\\
\midrule
Input & $\vh_c$ or $\vh_u$ & $8\times 8\times 384$ \\
Shake Shake dec layer & $4$ blocks, $384$ filters, $3\times3$,  stride 2 & $8\times 8\times 192$ \\
Shake Shake dec layer & $4$ blocks, $192$ filters, $3\times3$,  stride 2 & $16\times 16\times 96$ \\
Shake Shake dec layer & $4$ blocks, $96$ filters, $3\times3$,  stride 1 & $32\times 32\times 16$ \\
Output & $\vxh_c$ or $\vxh_u$ & $32\times 32 \times 3$ \\
\bottomrule
\end{tabular}
\begin{tablenotes}
\end{tablenotes}
\end{threeparttable}
\end{table}

\begin{table}[htbp]
\centering
\caption{Evolution of weights for HybridNet ResNet architecture for CIFAR-10}
\label{tab:resnetsched}
\begin{threeparttable}
\setlength{\tabcolsep}{4pt}
\begin{tabular}{ l l l}
\toprule
 & Value & Scheduling \\
\midrule
$\eta$ & 0.04 & Cosine decrease over the full training \\
$\lambda_c$ & 1 & Constant \\
$\lambda_s$ & 300 & Constant \\
$\lambda_r$ & 0.25 & Exponential increase over the first 5 epochs \\
${\lambda_r}_{b,l}$ & 0.5 & Exponential increase over the first 2 epochs \\
\bottomrule
\end{tabular}
\begin{tablenotes}
$\eta$ is the learning rate, $\lambda_c$ the classification weight, $\lambda_s$ the stability weight, $\lambda_r$ the final reconstruction weight, ${\lambda_r}_{b,l}$ the intermediate reconstructions weights.\\
Exponential decrease follows the function $\exp(-5t^2)$ with $t\in[0,1]$ from the start to the end of the decreasing interval. When increasing, $t$ goes from 1 to 0.\\
Cosine decrease follows the function $\cos(\pi t)+1$ with $t\in[0,1]$ from the start to the end of the decreasing interval.
\end{tablenotes}
\end{threeparttable}
\end{table}

\begin{table}[htbp]
\centering
\caption{Evolution of weights for HybridNet ResNet architecture for SVHN}
\label{tab:resnetschedsvhn}
\begin{threeparttable}
\setlength{\tabcolsep}{4pt}
\begin{tabular}{ l l l}
\toprule
 & Value & Scheduling \\
\midrule
$\eta$ & 0.04 & Cosine decrease over the full training \\
$\lambda_c$ & 1 & Constant \\
$\lambda_s$ & 100 & Exponential increase over the first 5 epochs \\
$\lambda_r$ & 0.1 & Exponential increase over the first 5 epochs \\
${\lambda_r}_{b,l}$ & 0.2 & Exponential increase over the first 2 epochs \\
\bottomrule
\end{tabular}
\begin{tablenotes}
$\eta$ is the learning rate, $\lambda_c$ the classification weight, $\lambda_s$ the stability weight, $\lambda_r$ the final reconstruction weight, ${\lambda_r}_{b,l}$ the intermediate reconstructions weights.\\
Exponential decrease follows the function $\exp(-5t^2)$ with $t\in[0,1]$ from the start to the end of the decreasing interval. When increasing, $t$ goes from 1 to 0.\\
Cosine decrease follows the function $\cos(\pi t)+1$ with $t\in[0,1]$ from the start to the end of the decreasing interval.
\end{tablenotes}
\end{threeparttable}
\end{table}

\begin{table}[htbp]
\centering
\caption{Evolution of weights for HybridNet ResNet architecture for STL-10}
\label{tab:resnetschedstl}
\begin{threeparttable}
\setlength{\tabcolsep}{4pt}
\begin{tabular}{ l l l}
\toprule
 & Value & Scheduling \\
\midrule
$\eta$ & 0.01 & Exponential decrease during the last 8 epochs \\
$\beta_1$ & 0.9 & Exponential increase during the last 80 epochs down to 0.5 \\
$\lambda_c$ & 0.1 & Constant \\
$\lambda_s$ & 0.1 & Exponential increase over the first 150 epochs \\
$\lambda_r$ & 0.01 & Exponential decrease during the last 17 epochs \\
${\lambda_r}_{b,l}$ & 0.01 & Exponential decrease during the last 17 epochs \\
\bottomrule
\end{tabular}
\begin{tablenotes}
$\eta$ is the learning rate, $\beta_1$ the first momentum of Adam, $\lambda_c$ the classification weight, $\lambda_s$ the stability weight, $\lambda_r$ the final reconstruction weight, ${\lambda_r}_{b,l}$ the intermediate reconstructions weights.\\
Exponential decrease follows the function $\exp(-5t^2)$ with $t\in[0,1]$ from the start to the end of the decreasing interval. When increasing, $t$ goes from 1 to 0.\\
Cosine decrease follows the function $\cos(\pi t)+1$ with $t\in[0,1]$ from the start to the end of the decreasing interval.
\end{tablenotes}
\end{threeparttable}
\end{table}


\end{document}